\documentclass[10pt,twocolumn,letterpaper]{article}

\usepackage{iccv}
\usepackage{times}
\usepackage{epsfig}
\usepackage{graphicx}
\usepackage{amsmath}
\usepackage{amssymb}
\usepackage{xcolor} 
\usepackage{adjustbox}
\usepackage{multirow}

\usepackage[breaklinks=true,bookmarks=false]{hyperref}

\iccvfinalcopy 

\def\iccvPaperID{10314} 

\ificcvfinal\fi

\begin{document}

\title{StorySync: Training-Free Subject Consistency in Text-to-Image Generation via Region Harmonization}

\author{Gopalji Gaur\\
University of Freiburg\\
{\tt\small gopaljigaur@gmail.com}
\and
Mohammadreza Zolfaghari\\
Zebracat AI\\
{\tt\small reza@zebracat.ai}
\and
Thomas Brox\\
University of Freiburg\\
{\tt\small brox@cs.uni-freiburg.de}
}

\maketitle

\ificcvfinal\thispagestyle{empty}\fi

\begin{figure*}[!ht]
    \centering
    \includegraphics[width=1\linewidth]{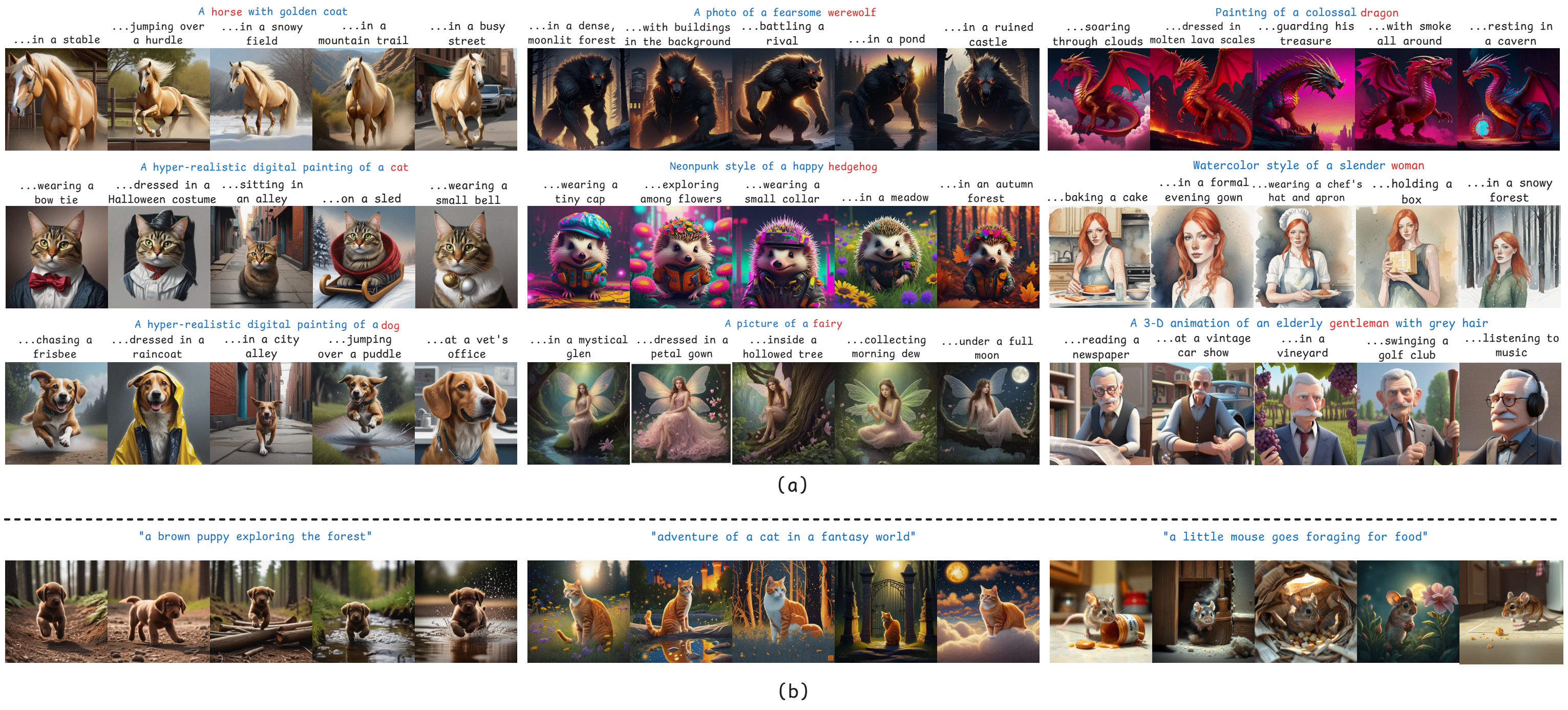}
    \caption[main figure]{\textbf{Demonstrating \textit{StorySync}'s consistency capabilities.} (a) Subject consistency maintained across various subject categories including humans, animals, and fictional characters. (b) Visual story generation showing StorySync's ability to maintain subject identity throughout multi-scene visual story sequences.}
    \label{fig:introduction}
\end{figure*}

\begin{abstract}

Generating a coherent sequence of images that tells a visual story, using text-to-image diffusion models, often faces the critical challenge of maintaining subject consistency across all story scenes. Existing approaches, which typically rely on fine-tuning or retraining models, are computationally expensive, time-consuming, and often interfere with the model’s pre-existing capabilities. In this paper, we follow a training-free approach and propose an efficient consistent-subject-generation method. This approach works seamlessly with pre-trained diffusion models by introducing \emph{masked cross-image attention sharing} to dynamically align subject features across a batch of images, and \emph{Regional Feature Harmonization} to refine visually similar details for improved subject consistency. Experimental results demonstrate that our approach successfully generates visually consistent subjects across a variety of scenarios while maintaining the creative abilities of the diffusion model.
\end{abstract}

\section{Introduction}

Current text-to-image diffusion models \cite{podell2023sdxlimprovinglatentdiffusion, arkhipkin2024kandinsky30technicalreport, rombach2022highresolution} struggle with maintaining subject consistency when generating multiple images. The lack of subject consistency in visual story generation extends beyond storytelling applications. In fields such as animation, game design, video creation, and synthetic data creation, consistent character representations are crucial for achieving coherence and realism. 

Various recent research efforts have explored methods to address this problem.
Most methods follow the idea of checkpoint personalization~\cite{Ruiz2022DreamBoothFT, dong_dreamartist_2023, kumari_multi-concept_2023, Avrahami_2024, liu2024intelligentgrimmopenended}, where the model is finetuned to generate a consistent subject. However, these approaches typically require extensive subject-specific training and struggle with incorporating multiple subjects in the same image \cite{Ruiz2022DreamBoothFT}. Other finetuning-based story generation approaches finetune either components of the diffusion model or an external layer to learn to generate consistent subjects \cite{wang2024characterfactorysamplingconsistentcharacters, he2025anystoryunifiedsinglemultiple, zhou2024storymaker}. However, all these approaches require additional training steps and suitable training data. In contrast, training-free methods, such as the IP-Adapter \cite{ye2023ipadaptertextcompatibleimage} use an image-conditioned diffusion process that takes a reference image as input and generates similar output images. Encoder-based approaches \cite{Arar2023DomainAgnosticTF, gal_encoder-based_2023, Li2023BLIPDiffusionPS, esser2021taming} focus on aligning the output image with a reference target, which restricts the model’s creative potential and produces images that do not closely follow the input prompts, therefore limiting creativity and preventing characters from adapting dynamically to new scenes.

Existing training-free consistent subject generation approaches \cite{tewel2024trainingfreeconsistenttexttoimagegeneration, zhou2024storydiffusionconsistentselfattentionlongrange} share visual knowledge about the subject among all images through attention sharing. Instead of relying on personalization or reference-based alignment, they leverage cross-image feature sharing to enforce zero-shot subject consistency in a batch of generated images. Other methods implement additional modifications such as text embedding weighting \cite{onepromptonestory}, or embedding clustering to generate visually similar subjects \cite{Avrahami_2024}. Despite achieving impressive subject consistency, these methods either fail to adhere to the conditioning prompts, or they lack alignment of finer details of the consistent subjects.

In this work, we introduce  \textit{StorySync}, which is built on \textbf{three technical innovations}. \textbf{(1)} We introduce masked cross-image attention sharing, a dense interaction of attention features localized to subject regions in the images. \textbf{(2)} To improve on the consistency of subtle subject details, we introduce Regional Feature Harmonization. \textbf{(3)} We present Base Layout Interpolation to enable sufficient diversity in the generated images, despite the consistency constraints.  As a result, \textit{StorySync} is able to generate story scenes with a high level of visual consistency of the story characters and surpasses existing training-free approaches, as shown in Figure \ref{fig:introduction}.

Furthermore, we demonstrate the plug-and-play capability of this approach on different text-to-image diffusion models. We test the approach with  \textit{SDXL} \cite{podell2023sdxlimprovinglatentdiffusion} and \textit{Kandinsky 3 }\cite{arkhipkin2024kandinsky30technicalreport}, both of which build on U-Net \cite{DBLP:journals/corr/RonnebergerFB15}, as well as with the time-distilled \textit{FLUX.1-schnell} model based on a transformer model.

\section{Related Work}

\textbf{Encoder-based approaches} \quad There have been multiple approaches \cite{jeong2023zeroshotgenerationcoherentstorybook, Su2023MakeAStoryboardAG, Pan2022SynthesizingCS,Zheng2024ContextualStoryCV, gong2023talecrafterinteractivestoryvisualization} to solve the challenge of generating visual stories using Diffusion Models \cite{ho2020denoisingdiffusionprobabilisticmodels}.
These approaches collectively address critical challenges such as maintaining coherence across scenes and ensuring character consistency in the generated scenes. Ye et al.~\cite{ye2023ipadaptertextcompatibleimage} introduced \textit{IP-Adapter}, a text-compatible image prompt adapter to enhance alignment between text and visuals, while Wei et al.~\cite{Wei2023ELITEEV} proposed \textit{ELITE}, which encodes visual concepts into textual embeddings for customized T2I generation. Jeong \etal \cite{jeong2023zeroshotgenerationcoherentstorybook} introduced a zero-shot framework leveraging Latent Diffusion Models and textual inversion to generate coherent storybooks directly from textual inputs. Building on this work, other approaches introduced frameworks for disentanglement of character and scene generation \cite{Su2023MakeAStoryboardAG}, or using a history-aware auto-regressive latent diffusion model \cite{Pan2022SynthesizingCS} to produce cohesive visual narratives. These contributions, although successful in generating cohesive and consistent storyboards, depend largely on input reference images to influence the storyboard generation.

\textbf{Model Finetuning} \quad Subsequent works have focused on model personalization to generate consistent subjects in diffusion models~\cite{zhu_domainstudio_2023, dong_dreamartist_2023, gal_encoder-based_2023, huang_reversion_2023, zhang_inversion-based_2023, Gal2022AnII, Richardson2023ConceptLabCG}. For example, Richardson et al.~\cite{Richardson2023ConceptLabCG} embedded novel concepts into a model's knowledge space, enabling consistent concept generation, while Arar et al.~\cite{Arar2023DomainAgnosticTF} fine-tuned models with real-world concepts in just 12 steps.

Key approaches like \textit{Textual Inversion}~\cite{gal_image_2022}, \textit{Dreambooth}~\cite{Ruiz2022DreamBoothFT}, and \textit{Custom Diffusion}~\cite{kumari_multi-concept_2023} associate new concepts with unique tokens in the text encoder dictionary, allowing models to reconstruct these concepts during generation. Gong et al.~\cite{gong2023talecrafterinteractivestoryvisualization} developed \textit{TaleCrafter} for interactive visual storytelling using customized \textit{LoRA}~\cite{hu_lora_2021} models, while Yang et al.~\cite{yang2024seedstorymultimodallongstory} trained multi-modal \textit{LoRA} models for consistent subject generation in long stories.

Fine-tuning T2I models on storyboard datasets has also shown promise for visual storytelling~\cite{feng2023improvedvisualstorygeneration, Liu2023IntelligentG}. Wu et al.~\cite{wu2023tuneavideooneshottuningimage} extended this approach to video generation, achieving temporal consistency with personalized T2I models. Decentralized methods~\cite{Gu2023MixofShowDL, Po2023OrthogonalAF} utilize ensemble models to generate multi-subject consistent scenes. Wang et al.~\cite{wang2024oneactorconsistentcharactergeneration} further advanced this by sampling latent noise from localized regions of the latent space, enabling consistent character generation. However, these approaches require computationally expensive fine-tuning of model checkpoints, limiting practical applications.

In addition to addressing coherence, multi-modal frameworks such as \textit{SEED-Story} by Yang et al. \cite{yang2024seedstorymultimodallongstory}, \textit{TaleCrafter} by Gong et al. \cite{gong2023talecrafterinteractivestoryvisualization} and Liu et al.'s \textit{Intelligent Grimm} \cite{liu2024intelligentgrimmopenended} push the boundaries of interactive and multi-modal storytelling using Latent Diffusion Models \cite{rombach2022highresolution}. Together, these works offer robust solutions for generating visually similar subjects, but often require fine-tuning the diffusion models to generate visually consistent subjects.

\begin{figure*}[ht]
    \centering
    \includegraphics[width=1\linewidth]{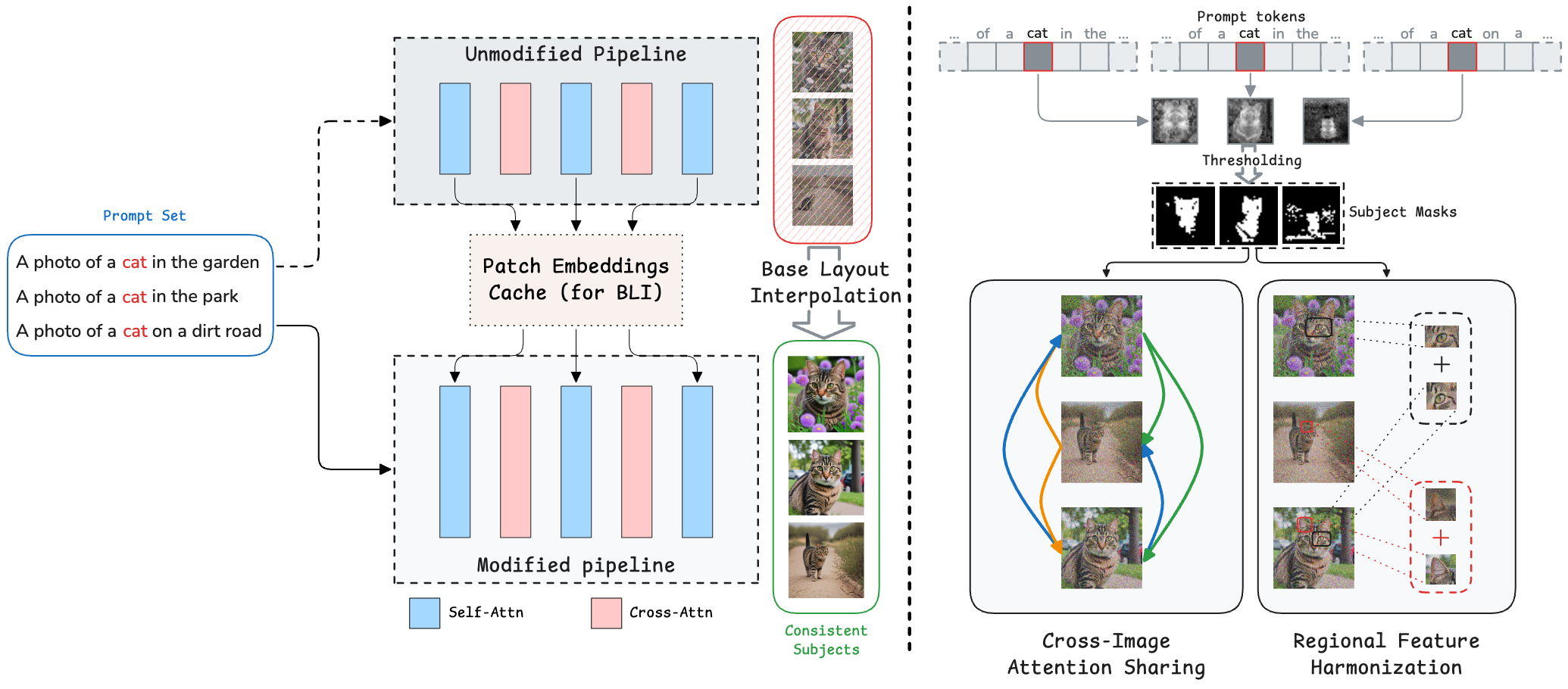}
    \caption[Cross-Image Attention Sharing]{\textbf{Proposed Architecture:} (1) We cache queries generated by the base model during image generation, (2) We modify the image generation pipeline's attention processors with our own implementations, (3) We generate images with consistent subjects using the modified pipeline. \textit{In modified Cross Attention layers}: we extract attention maps to generate subject masks. \textit{In modified Self-Attention layers}: We implement Cross-Image Attention Sharing and Regional Feature Harmonization to enforce subject consistency in generated images.}
    \label{fig:complete-approach}
\end{figure*}

\textbf{Training-free Consistency} \quad Achieving one-shot consistency in generated images is crucial to enabling visually coherent stories or images without the overhead of fine-tuning or additional computational resources~\cite{tewel2024trainingfreeconsistenttexttoimagegeneration, zhou2024storydiffusionconsistentselfattentionlongrange}. Shi et al.~\cite{Shi2023InstantBoothPT} introduced \textit{InstantBooth}, offering near-instant model personalization without test-time fine-tuning. In image editing, Cao et al.~\cite{Cao2023MasaCtrlTM} proposed \textit{MasaCtrl}, which employs mutual self-attention to share information between input and generated images. Wang et al.~\cite{wang2024spotactortrainingfreelayoutcontrolledconsistent} introduced \textit{RISA} and \textit{SFCA} mechanisms to enforce layout-defined subject consistency.

Training-free approaches such as \textit{ConsiStory} \cite{tewel2024trainingfreeconsistenttexttoimagegeneration} and \textit{StoryDiffusion} \cite{zhou2024storydiffusionconsistentselfattentionlongrange} introduce self-attention modifications to enforce subject consistency. He et al.~\cite{he2024dreamstoryopendomainstoryvisualization} proposed \textit{DreamStory}, focusing on open-domain story visualization with attention sharing among images. ConsiStory enforces strong cross-frame context and query blending, which suppresses pose diversity. StoryDiffusion propagates context beyond subject regions, resulting in repeated visual patches. In contrast, our method injects pose cues via an independent branch and constrains attention using subject masks, improving both consistency and diversity. We take inspiration from such techniques and devise our Cross-Image Attention Sharing, Regional Feature Harmonization, and Base Layout Interpolation to generate compelling visual stories with consistent subjects.  

\section{StorySync}

In this section, we present our approach \textit{StorySync} for achieving subject consistency in Text-to-Image generation pipelines.  
As shown in Figure \ref{fig:complete-approach}, StorySync enhances subject consistency through three primary mechanisms: (1) Cross-Image Attention Sharing restricted to subject regions, (2) Regional Feature Harmonization to strengthen fine-grained visual similarity of subject across generated images, and (3) Base
Layout Interpolation to boost prompt adherence. Together, these mechanisms enforce subject consistency while preserving scene diversity.

\subsection{Preliminaries}

\textbf{Extracting QKV tensors} \quad Given an input sequence of image patch embeddings $X \in \mathbb{R}^{N \times d}$, where $N$ is the number of patch tokens and $d$ is the embedding dimension, each token in $X$ is linearly transformed to produce queries ($Q$), keys ($K$), and values ($V$):
    \begin{multline} \label{eq:qkv}
        Q = XW_Q, \quad K = XW_K, \quad V = XW_V, \\ W_Q, W_K, W_V \in \mathbb{R}^{d \times d_k},
    \end{multline}
    where $d_k$ is the dimension of the projected space.
These tensors are then used during self-attention computation.

\textbf{Extracting Subject Masks} \quad To ensure consistency only in subject regions, we extract cross-attention maps associated with the subject token in Cross-Attention layers of the de-noising network.

For a given subject token $s$ in the textual input, the attention map for an image \(i\) is computed as:
\begin{align}
    \mathcal{A}_{s,i} = \text{softmax}\left(\frac{Q_{s,i} K_{i}^T}{\sqrt{d_k}}\right),
\end{align}
where $Q_{s,i}$ represents the query vector corresponding to the subject token, and $K_{i}$ are the key vectors derived from the patch embeddings of image \(i\). 

The resulting attention map $\mathcal{A}_{s,i}$ is averaged over cross-attention layers in the model, and then summed for all subjects \(S\) in an image \(i\) to create a robust representation of the influence of the subjects in image patches:
\begin{align}\label{eq:masks}
    \bar{\mathcal{A}}_{s,i} &= \frac{1}{L} \sum_{l=1}^L \mathcal{A}_{s,i}^{(l)}, \\
    \bar{\mathcal{A}_i} &= \sum_{s=1}^S \bar{\mathcal{A}}_{s,i}
\end{align}
where $L$ is the total number of layers from which the maps were extracted, and \(\bar{\mathcal{A}_i}\) is the aggregated map.

Finally, thresholding with Otsu's method~\cite{Otsu1979ATS} converts the attention maps \(\bar{\mathcal{A}_i}\) into binary subject masks \(\mathcal{M}_i\) for each image \(i\). In addition to using cross-attention maps for generating subject masks, we experimented with utilizing segmentation of the intermediate latents to obtain subject masks, more details about this experiment are included in Appendix \ref{appendix:segmentation-masks}.

\subsection{Boosting subject consistency}
A straightforward approach to promoting subject consistency across multiple generated images is to extend the self-attention mechanism, allowing queries from one image to attend to keys and values from other images in the batch, the approach followed in concurrent works \cite{tewel2024trainingfreeconsistenttexttoimagegeneration, zhou2024storydiffusionconsistentselfattentionlongrange}.

To prevent unwanted consistency in background patches across images, we use subject masks \cite{tewel2024trainingfreeconsistenttexttoimagegeneration}, obtained in Equation \ref{eq:masks}. Unlike ConsiStory~\cite{tewel2024addittrainingfreeobjectinsertion}, we aggregate cross-attention maps only from the current generation timestep to prevent overhead of storing the cross-attention maps from previous generation timesteps. This temporal optimization enables \textit{StorySync} to achieve superior subject consistency with less computational cost while it's more responsive to the evolving state of the generation process. 

\subsubsection{Cross-Image Attention Sharing}

\begin{figure}
    \includegraphics[width=1\linewidth]{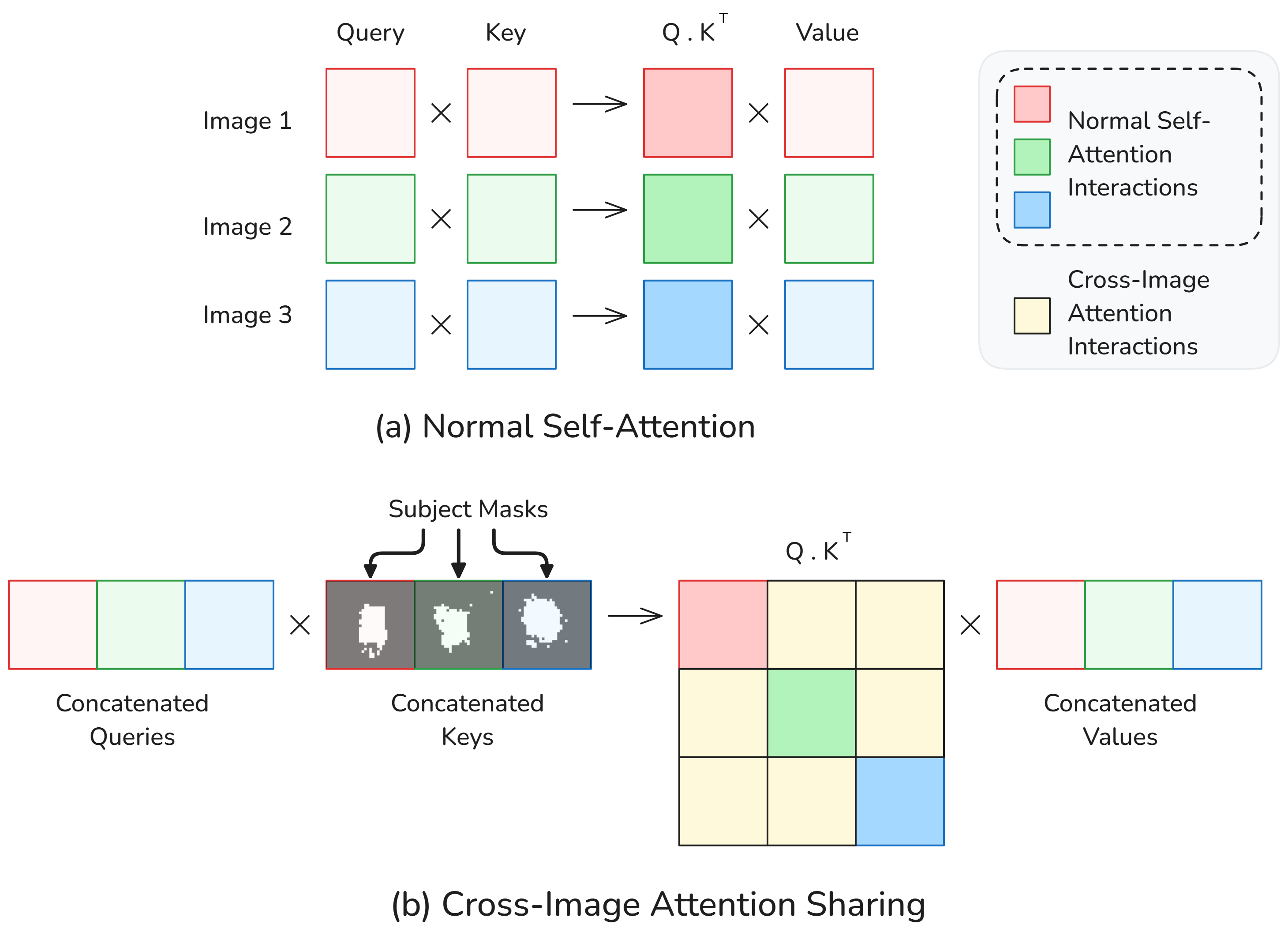}
    \caption[Cross-Image Attention Sharing]{\textbf{Cross-Image Attention Sharing.} Contrary to normal Attention Calculation (a), we enable interaction among the Query, Key, and Value tensors from subject regions across images (b)}
    \label{fig:cross-image-attn}
\end{figure}

Our Cross-Image Attention Sharing mechanism enables controlled interaction between patches in the current image and those sampled from subject regions across other images in the batch of size \(N\) (Figure~\ref{fig:cross-image-attn}). 
By utilizing the subject masks \(\mathcal{M}_i\) (see Equation~\ref{eq:masks}), we constrain attention calculation to ensure information flows exclusively between subject-specific regions across different images while preserving standard self-attention within each individual image.

To enforce these constraints, we define a propagation mask $\Gamma_i$, which determines which regions in other images can attend to the current image:
\begin{align}
\Gamma_i = \bigoplus_{j=1}^N \delta_{ij} \mathcal{M}_j, \quad \text{where} \quad \delta_{ij} =
\begin{cases}
\mathbf{1}, & j = i \\
\mathcal{M}_j, & j \neq i
\end{cases}.
\end{align}
Here, $\bigoplus$ denotes the concatenation operation between subject masks, and $\delta_{ij}$ ensures that the propagation masks includes both self and cross-image subject masks to preventing background interference while preserving self-image features during attention calculation.

Once the Query, Key, and Value tensors are obtained for each image in the batch (Equation~\ref{eq:qkv}), we stack the obtained matrices:
\begin{align}
Q_{\text{all}} &= \bigcup_{j=1}^{N} Q_j,  \quad
K_{\text{all}} = \bigcup_{j=1}^{N} K_j,  \quad
V_{\text{all}} = \bigcup_{j=1}^{N} V_j
\end{align}
Here, $\bigcup$ denotes a set-like stacking operation that preserves individual image structures but allows joint computations across images.

The attention mechanism then becomes:
\begin{align}
\mathcal{A}_i &= \text{softmax} \left( \frac{Q_i (K_{\text{all}})^T}{\sqrt{d_k}} + \log \Gamma_i \right), \\
h_i &= \mathcal{A}_i \cdot V_{\text{all}}.
\end{align}

where $\mathcal{A}_i$ represents the attention matrix, where regions with $\Gamma_i = 0$ are assigned $-\infty$ before applying softmax, ensuring restricted interactions and $h_i$ is the output activation incorporating information from relevant subject regions across all images in the batch.

\subsubsection{Regional Feature Harmonization}
Fine-grained visual alignment of subject-specific attributes (e.g. facial details, chromatic consistency, or textural patterns) is fundamentally challenging in multi-image story generation, particularly when attempting to enforce both identity consistency and contextual variation.

We propose Regional Feature Harmonization (RFH) based on the principle that semantically equivalent regions across generated images should maintain visual coherence while preserving their contextual uniqueness. 
We use a distribution-based correspondence approach that identifies and aligns similar regions across the image batch. Unlike static feature injection approaches~\cite{tewel2024trainingfreeconsistenttexttoimagegeneration} that rely on pre-computed DIFT embeddings~\cite{Tang2023EmergentCF}, our method calculates feature alignments in real-time during de-noising iterations, as visualized in Figure~\ref{fig:complete-approach}-right.

RFH utilizes intermediary region representations $\mathcal{R}_i$ from self-attention block, which capture rich textural and structural information. To identify optimal region correspondences between images $i$ ($I_i$) and $j$ ($I_j$), we formulate a region-wise compatibility function:

\begin{align}
\mathcal{H}_{i,j}(r,\omega) = \frac{\exp(\langle\mathcal{R}_i(r), \mathcal{R}j(\omega)\rangle / \tau)}{\sum_{\omega' \in \Omega_j} \exp(\langle\mathcal{R}_i(r), \mathcal{R}_j(\omega')\rangle / \tau)}
\end{align}
where $\langle\cdot,\cdot\rangle$ denotes normalized inner product, and $\Omega_j$ is defined as the index set of foreground patches belonging to an image $I_j$, and $\tau$ is a temperature parameter.
The optimal region mapping function $\mathcal{C}_{i}(r, I_j)$ for the region $r$ of $I_i$ and all regions of $I_j$ is then obtained:

\begin{align}
\mathcal{C}_{i}(r, I_j) = { \omega^* \quad \text{where} \quad \omega^* = \underset{\omega \in \Omega_j}{\mathrm{argmax}} , \mathcal{H}_{i,j}(r, \omega), j \neq i}
\end{align}

The corresponding regions are then harmonized through an adaptive regional fusion mechanism:
\begin{align}
\hat{\mathcal{R}}_i(r) = \mathcal{R}_i(r) + \gamma \cdot \mathcal{M}_i(r) \cdot (\mathcal{R}_j(\mathcal{C}_i(r, I_j)) - \mathcal{R}_i(r))
\end{align}

where $\gamma$ represents the harmonization coefficient. The term $(\mathcal{R}_j(\mathcal{C}_i(r, I_j)) - \mathcal{R}_i(r))$ represents the feature adaptation vector needed to transform region $r$'s features to more closely match its correspondence in image $I_j$. By adding a scaled version of this difference vector to the original region features, we're effectively pushing the region's representation toward its correspondence in feature space. 
With subject mask $\mathcal{M}_i(r)$, we restrict harmonization to subject regions to keep background features unaffected. Additionally, Otsu’s Thresholding~\cite{Otsu1979ATS} limits harmonization only to regions with sufficiently high correspondence.

\subsection{Boosting Prompt Adherence}

Since early generation steps heavily influence layout formation~\cite{Patashnik2023LocalizingOS, balaji2023ediffitexttoimagediffusionmodels}, some works disable consistent subject generation during these steps to allow pose variation at the cost of subject similarity~\cite{zhou2024storydiffusionconsistentselfattentionlongrange}.

ConsiStory~\cite{tewel2024trainingfreeconsistenttexttoimagegeneration} performs vanilla query blending to incorporate vanilla subject poses, however, the pose diversity is diluted due to the integration of query blending steps directly in image generation timesteps. We introduce a Base Layout Interpolation (BLI) method to incorporate the pose information from the images generated by the base model into our generated images.

\begin{figure*}[ht]
    \centering
    \includegraphics[width=1\linewidth]{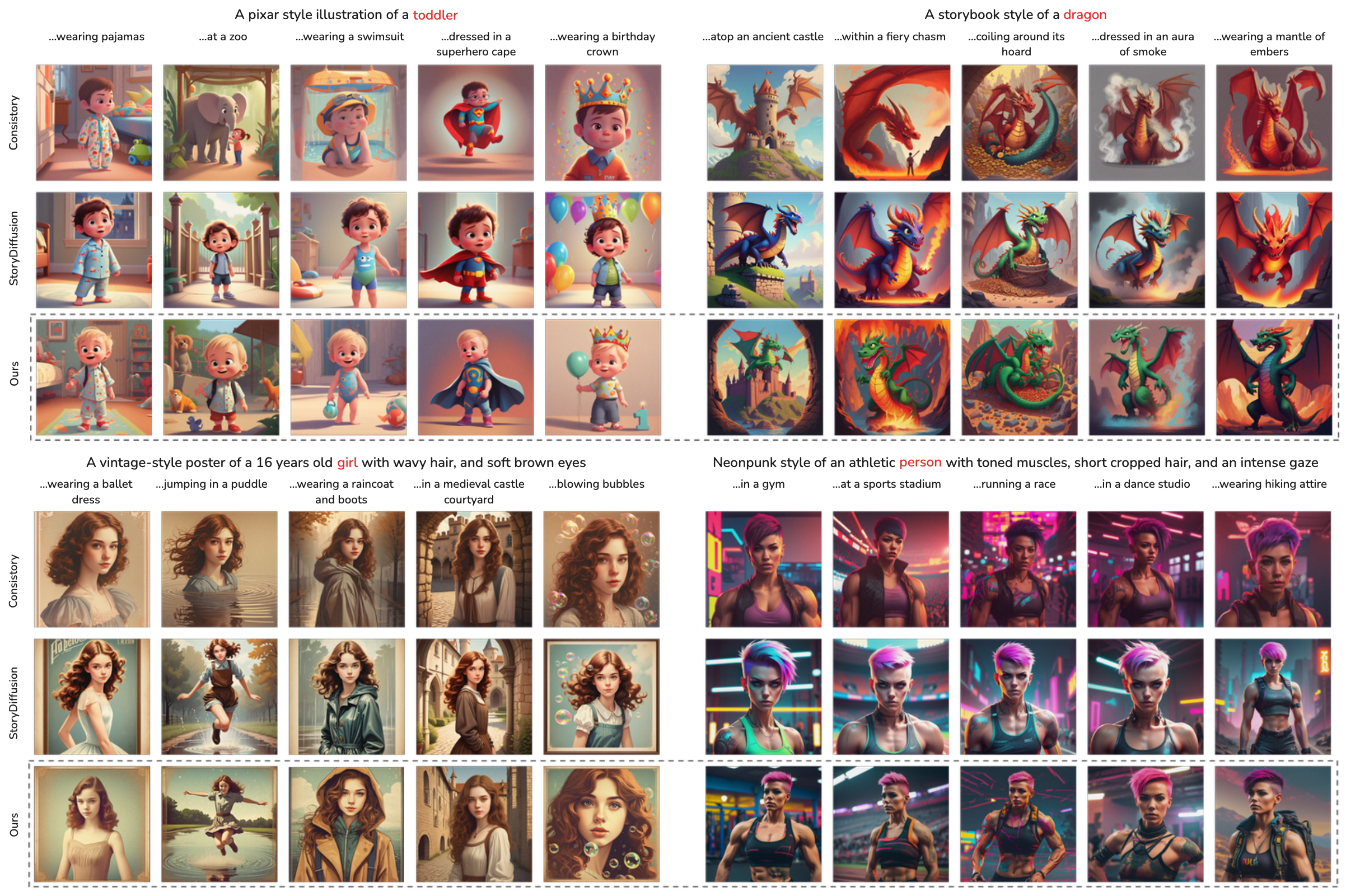}
    \caption[Qualitative Results]{\textbf{Qualitative Results.} While ConsiStory \cite{tewel2024trainingfreeconsistenttexttoimagegeneration} achieves subject consistency, it sometimes has problems adhering to subject prompts (e.g. in \textit{girl} and \textit{person} examples, it generates same poses irrespective of prompt), StoryDiffusion \cite{zhou2024storydiffusionconsistentselfattentionlongrange} struggles with subject consistency (e.g. in \textit{dragon} example) in some cases because it does not use subject masking. Our approach achieves overall excellent subject consistency while also following the instructions provided by the input prompts.}
    \label{fig:comparison_consistory}
\end{figure*}

BLI is implemented in two steps and it helps \textit{StorySync} achieve high level of prompt adherence. Initially, we start image generation using the vanilla base model and at each timestep \(t\), and in each self-attention layer \(l\) of the model, we cache the intermediate patch embeddings \(X_{t,cached}^l\). These embeddings capture rich compositional information from the prompt-driven, unconstrained generation process. Next, we start the consistency-enhanced image generation process, and during each timestep \(t\), the patch embeddings in each self-attention layer \(X_{t,consist}^l\) are adaptively integrated with the compositional guidance from the cached embeddings:

\begin{align}
\begin{split}
\text{Step 1:} \quad & \text{Vanilla denoising and cache embeddings} \\
& X_{t,\text{cached}}^l \leftarrow \text{Denoise}_{\text{vanilla}}(z_t, l) \quad \forall t \in T, \forall l \in L \\
\text{Step 2:} \quad & \text{Consistency-enhanced denoising with alignment} \\
& X_{t,\text{consist}}^l \leftarrow \text{Denoise}_{\text{consistent}}(z_t, l) \\
& X_{t,\text{final}}^l \leftarrow (1-\lambda) \cdot X_{t,\text{consist}}^l + \lambda \cdot X_{t,\text{cached}}^l
\end{split}
\end{align}

where \( \lambda \) controls the degree of interpolation, and \(T\) is the number of timesteps we perform BLI for. 

As shown in Figure~\ref{fig:complete-approach}, we are able adapt the poses and layouts of the subjects to their layouts as generated by the base model. By decoupling the embedding caching step from the consistency-enhanced generation step, we incorporate diverse compositional information that might otherwise be homogenized by consistency mechanisms. To further enhance prompt adherence, we introduce dropouts in the different sections of our approach that boost subject consistency: 1) Cross-Image Attention Sharing, 2) Regional Feature Harmonization, and 3) in the subject masks.

\subsection{Scalable image generation.}
To efficiently handle subject consistency across large image batches, we optimize the generation process by initially generating a subset of images. This technique aligns with methods used in recent works~\cite{tewel2024trainingfreeconsistenttexttoimagegeneration, zhou2024storydiffusionconsistentselfattentionlongrange, wang2024oneactorconsistentcharactergeneration}. These initially generated images serve as primary reference sources for cross-image interactions during both Cross-Image Attention Sharing and Regional Feature Harmonization. This technique allows us to generate longer sequences while reducing computational cost and maintaining consistency.

For a batch containing two subset images, we redefine the Key and Value matrices as follows:
\begin{align}
K_{\text{sub}} = \bigcup_{j \in \{1,2\}} K_j, \quad
V_{\text{sub}} = \bigcup_{j \in \{1,2\}} V_j.
\end{align}
Here, $K_{\text{sub}}$ and $V_{\text{sub}}$ exclusively contain information from the subset images. Queries $Q_i$, where \(i \in \{1, 2, \dots, N\} \setminus \{1, 2\}\), interact solely with these matrices during attention computation. These subset tensors can be cached and reused in future image generation processes.

\section{Experiments}

\subsection{Qualitative Analysis}
In order to generate the qualitative results across a wide variety of subjects, we utilize prompts generated by Chat GPT \cite{Liu_2023} for different classes of subjects (Eg. \textit{dog}, \textit{old man}, etc.), in different settings, (\textit{on the road}, \textit{on the beach}, etc.), and image styles (\textit{realistic photo}, \textit{watercolor painting}, etc.). In addition to these prompts, we also utilize ConsiStory~\cite{tewel2024trainingfreeconsistenttexttoimagegeneration} benchmark prompts to compare our results with other approaches.

\begin{table*}[ht]
\begin{center}
    \begin{tabular*}{\textwidth}{@{\extracolsep{\fill}} l c c c c c }
        \hline
        Method       & Base Model & CLIP-T\(\uparrow\) & CLIP-I\(\uparrow\) & LPIPS\(\uparrow\) & DreamSim\(\downarrow\) \\ \hline
        Base SDXL & - & 0.8749 & 0.7819 & 0.3497 & 0.5263 \\ 
        Base Kandinsky 3 & - &  0.8758 & 0.7944 & 0.3239 & 0.4929 \\
        Base FLUX.1 & - & 0.8968 & 0.8026 & 0.3320 & 0.4806 \\ \hline
        ConsiStory & SDXL & 0.8071 & 0.8289 & 0.3996 & 0.3440 \\
        StoryDiffusion & SDXL & \textbf{0.8126} & 0.8572 & \textbf{0.4198} & 0.3589 \\
        \textit{StorySync (Ours)} & SDXL & 0.8108 & \textbf{0.8735} & 0.4143 & \textbf{0.2869} \\ \hline
        \textit{StorySync (Ours)} & Kandinsky 3 & 0.8075 & 0.8763 & 0.4091 & 0.2804 \\
        \textit{StorySync (Ours)} & FLUX.1-schnell & 0.8244 & 0.8765 & 0.4039 & 0.2883 \\
        \hline
    \end{tabular*}
    \end{center}

    \caption[Quantitative Analysis]{\textbf{Quantitative Analysis.} We compare the performance of our approach against Consistory \cite{tewel2024trainingfreeconsistenttexttoimagegeneration} and StoryDiffusion \cite{zhou2024storydiffusionconsistentselfattentionlongrange} on \textit{SDXL} model. The best score in each column is highlighted in \textbf{bold}. Our approach achieves better scores on both perceptual similarity metrics (CLIP-I, LPIPS, DreamSim) and prompt alignment metric (CLIP-T). Additionally we present the scores of our approach on \textit{Kandinsky 3}, and \textit{FLUX.1} models. Base model scores are for reference only. \\}
    \label{tab:quantitative}
\end{table*}

In Figure \ref{fig:comparison_consistory}, we present qualitative comparisons of our approach with state-of-the-art training-free approaches \cite{tewel2024trainingfreeconsistenttexttoimagegeneration, zhou2024storydiffusionconsistentselfattentionlongrange}. For comparison, for each prompt \(p\), we generate the output image using ConsiStory \cite{tewel2024trainingfreeconsistenttexttoimagegeneration}, StoryDiffusion \cite{zhou2024storydiffusionconsistentselfattentionlongrange}, and \textit{StorySync} on a pre-trained SDXL model. From the Figure \ref{fig:comparison_consistory}, we can observe that our approach achieves a high degree of subject consistency while maintaining strong adherence to the input text prompts. While ConsiStory \cite{tewel2024trainingfreeconsistenttexttoimagegeneration} is able to generate visually similar subjects, the alignment of the generated images is low, and, in the case of StoryDiffusion \cite{zhou2024storydiffusionconsistentselfattentionlongrange}, the subjects follow the input prompts but sometimes lack visual subject consistency. Our approach achieves the best balance between subject consistency and prompt adherence.

\begin{figure*}[ht]
    \centering
    \includegraphics[width=1\linewidth]{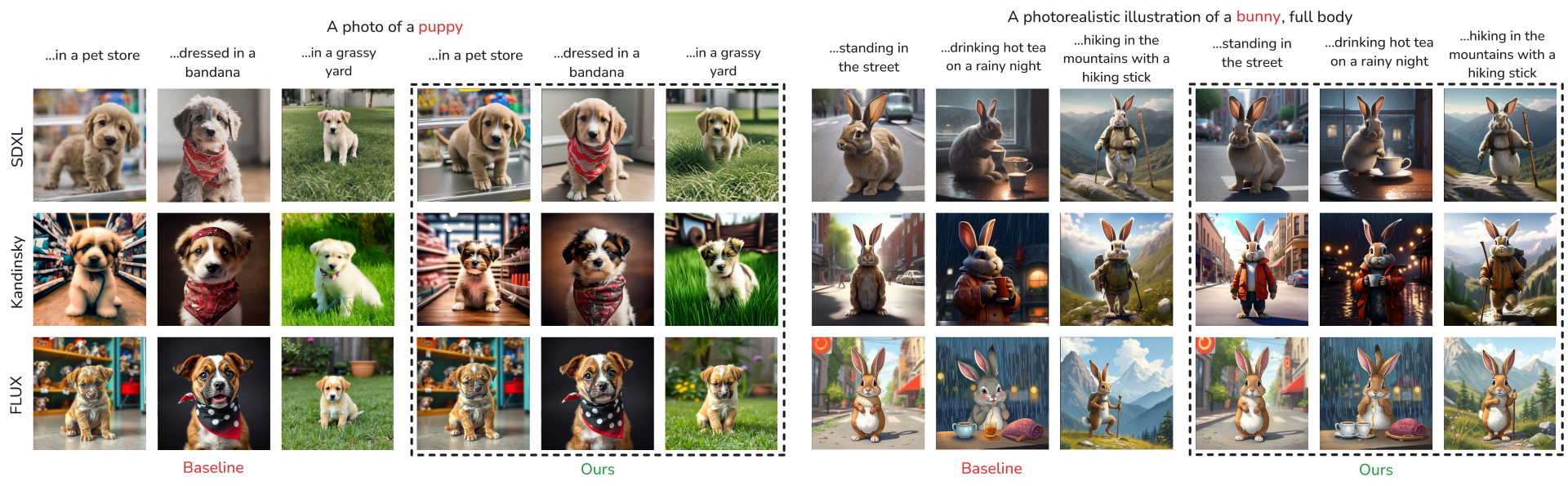}
    \caption[Consistent Subject Generation Examples]{\textbf{Results with multiple diffusion models.} We present the results when \textit{StorySync} is integrated with multiple T2I Diffusion Model pipelines (\textit{SDXL}, \textit{Kandinsky 3} and \textit{FLUX.1-schnell}). Our approach consistently generates visually similar subjects for each of these models.}
    \label{fig:consistent-sub-gen}
\end{figure*}

Furthermore, to the best of our knowledge, StorySync is one of the first training-free consistent subject generation techniques that works across different classes of text-to-image diffusion models. \textit{StorySync}'s simple design enables us to integrate it to multiple T2I models with simple hyperparameter tuned for each pipeline. In Figure \ref{fig:consistent-sub-gen}, we demonstrate \textit{StorySync}'s capability in generating consistent subjects using SDXL, Kandinsky 3 and FLUX.1-schnell models. The pose diversity appears limited in Figure \ref{fig:consistent-sub-gen}. This is due to FLUX-1’s use of Rotatory Positional Embeddings (RoPE), which encode strong spatial priors. When shared across images, this tends to align subject poses more closely. More results with further Qualitative Analysis of our approach can be found in Appendix \ref{appendix:complex-prompts-multiple-subjects}.

\subsection{Quantitative Analysis}

In this section, we evaluate \textit{StorySync} using quantitative metrics to evaluate the prompt adherence of the generated images and the visual similarity of characters in the images. We utilize the ConsiStory benchmark introduced by Tewel et al. \cite{tewel2024trainingfreeconsistenttexttoimagegeneration} for evaluations. To quantify both prompt adherence and subject similarity across the generated images, we employ \textit{CLIP-score} \cite{radford2021learning} as our primary evaluation metric. We denote the score for prompt adherence using CLIP embeddings as CLIP-T and for subject similarity as CLIP-I as used in previous similar works \cite{onepromptonestory}. We use Learned Perceptual Image Patch Similarity (LPIPS) \cite{zhang2018unreasonableeffectivenessdeepfeatures} and DreamSim \cite{fu2023dreamsimlearningnewdimensions} to measure the similarity of generated subjects in the images. Background of the images is removed using Carvekit to measure the perceptual similarity of only the subjects. It is to be noted that during evaluation \textit{DreamSim} scores should be given more preference , as they better align with human perceptual assessment of image similarity \cite{fu2023dreamsimlearningnewdimensions}. In contrast, LPIPS \cite{zhang2018unreasonableeffectivenessdeepfeatures} primarily quantifies visual similarity based on spatial layout.

In Table \ref{tab:quantitative}, we report the quantitative comparison of our approach \textit{StorySync}, against SOTA consistent subject generation approaches such as ConsiStory \cite{tewel2024trainingfreeconsistenttexttoimagegeneration} and StoryDiffusion \cite{zhou2024storydiffusionconsistentselfattentionlongrange}. From the Table \ref{tab:quantitative}, we can see that \textit{StorySync} achieves the best scores in CLIP-I and DreamSim metrics, compared to ConsiStory \cite{tewel2024trainingfreeconsistenttexttoimagegeneration}, and StoryDiffusion \cite{zhou2024storydiffusionconsistentselfattentionlongrange} when evaluated on SDXL model. StoryDiffusion \cite{zhou2024storydiffusionconsistentselfattentionlongrange} shows marginally higher on CLIP-T and LPIPS scores mainly due to its random masking approach during attention sharing, which induces feature averaging across subjects. 
However, this metric advantage does not necessarily translate to better visual coherence in generated subjects. It is important to mention that these scores alone do not fully reflect the method’s effectiveness in maintaining subject consistency or adhering to the given prompt.
Therefore, these quantitative results should be interpreted alongside qualitative evaluations, as emphasized by Tewel et al. \cite{tewel2024trainingfreeconsistenttexttoimagegeneration}. Human inspection of the visual quality of the images is important for measuring the ability our approach in generating consistent subjects. The results for the human evaluation study comparing the three approaches can be found in Appendix \ref{appendix:human-evaluation}.

\begin{figure}
\centering
    \includegraphics[width=1\linewidth]{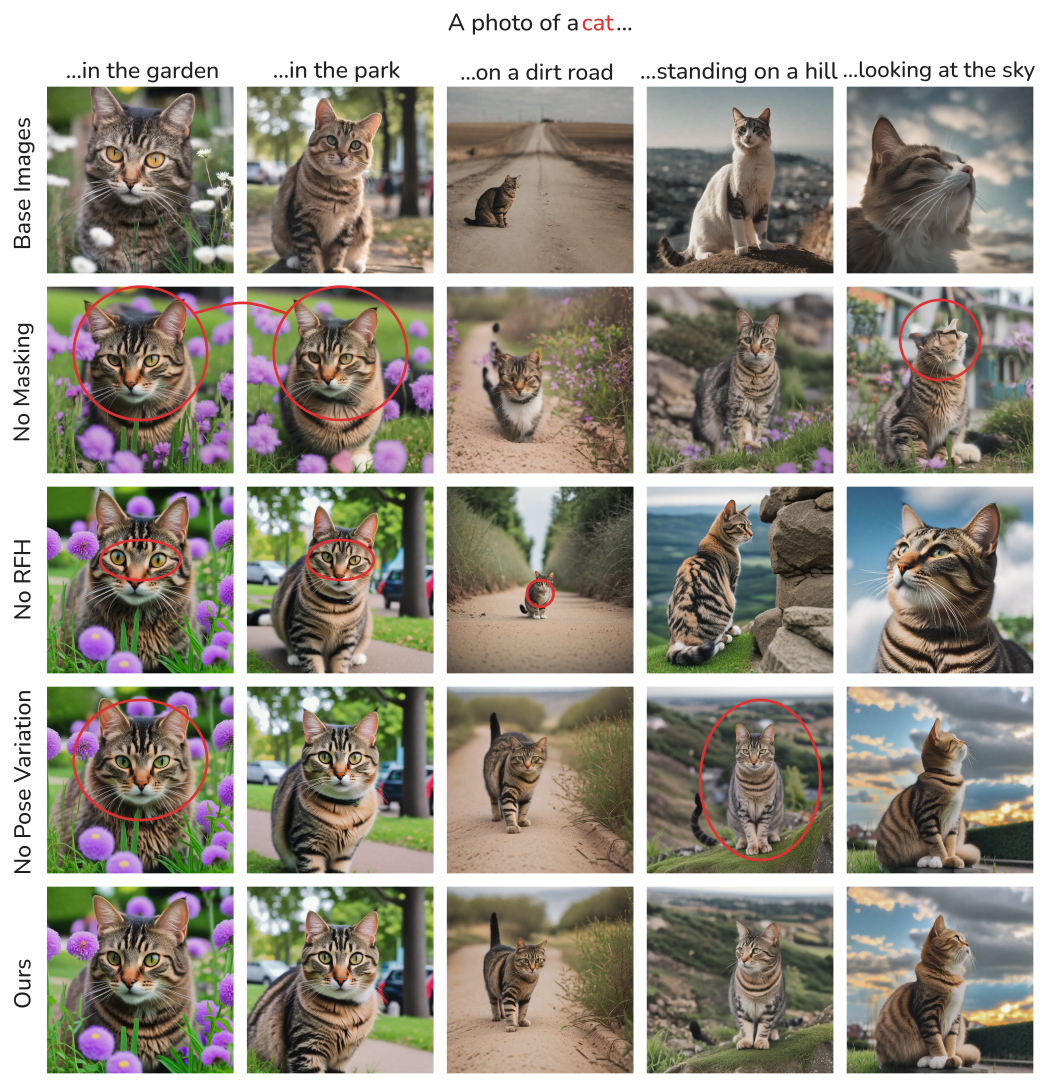}
    \caption[Ablation Study]{\textbf{Ablation Study.} Without using subject masks, we observe similarity in subject poses and some deformation (red circle in second row); Without RFH, smaller details of subject such as the eyes and coat pattern are less similar; Images without Pose variation techniques (BLI and Dropouts) have a cat that faces only forward in all images}
    \label{fig:ablation}
\end{figure}

\subsection{Ablation}
In this section, we inspect the effect of different components of \textit{StorySync} and study their effect on the overall image generation quality and consistent subject generation. We compare the effects of Subject Masks, RFH, and Pose variation (BLI, and dropouts) on the generated images. For this study, we disable a component of our approach one at a time, while keeping all the other components enabled. We generate 5 images of a simple subject (\textit{cat}), with short and simple prompts to prevent any interference in the experiment due to prompt complexity.

From Figure \ref{fig:ablation}, we can observe that without any of the components of our approach enabled, the subject \textit{cat} is visually dissimilar in all 5 generated images. When we disable the subject masks, the visual appearance and layout of the subject, as well as the backgrounds are identical in images. We also observe deformation of some subject regions due to uncontrolled RFH in background regions. Hence, subject masks play an important role in generating visually appealing subjects in a visual story with consistent subjects.

Without RFH, there is some deterioration in subject similarity (Figure \ref{fig:ablation}). Especially the image for the prompt "\textit{a photo of a cat on a dirt road}", in which the subject no longer resembles other subjects, shows this decline. When disabling pose variation techniques, we observe increased similarity in subject poses across the images, highlighting the importance of BLI and Dropouts in enforcing prompt adherence and layout diversity. To study the effects of the components of our approach in a quantitative manner, we have included the results of the qualitative ablation study in Appendix \ref{appendix:quant-ablation-study}.

\subsection{Limitations}

\begin{figure}
\centering
    \includegraphics[width=0.74\linewidth]{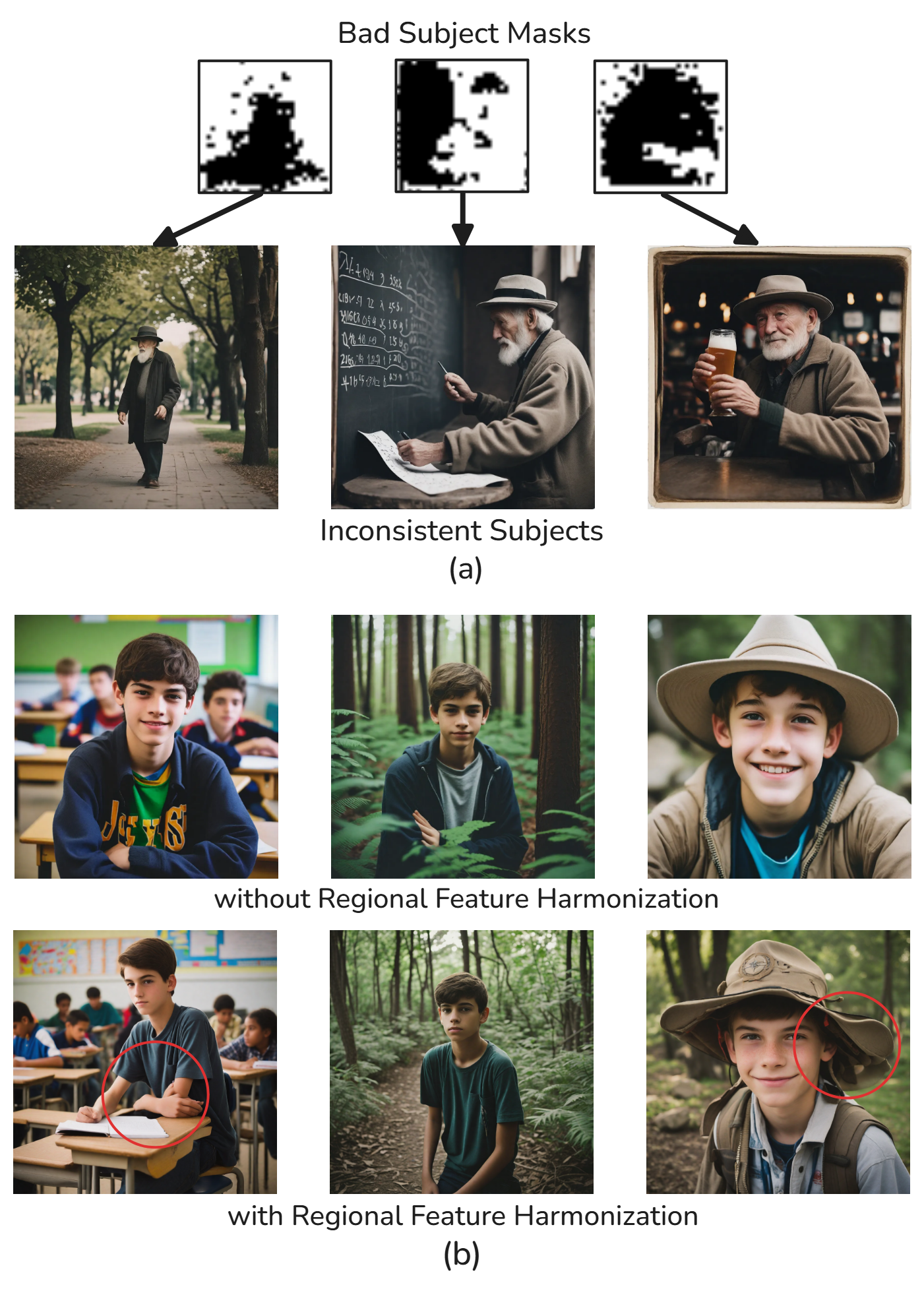}
    \caption[Limitations]{\textbf{Limitations:} (a) Incorrect subject masks can block attention sharing and therefore lead to inconsistency; (b) Subject deformation (red circle) may occur when unrelated regions are fused with the subject regions.}
    \label{fig:limitations}
\end{figure}

One limitation of \textit{StorySync} is its dependence on subject masks from cross-attention maps. If these masks fail to align with subject regions, it can cause inconsistencies in the rendered subject across different images (Figure \ref{fig:limitations}a). Additionally, Regional Feature Harmonization can occasionally misidentify corresponding regions based on color, texture, or pattern similarity. Such misalignment can deform the generated subject or introduce inconsistencies in fine details, as seen in Figure \ref{fig:limitations}b. However, these issues occur in only a small fraction of cases.

\section{Conclusion}

We introduced \textit{StorySync}, an approach for subject consistency in text-to-image diffusion models, a critical aspect for applications like visual storytelling, animation, and content creation. Building upon previous training-free consistency approaches, we developed a comprehensive pipeline that integrates three key components: Masked cross-Image Attention Sharing, Regional Feature Harmonization, and Base Layout Interpolation. Our extensive experiments demonstrate that \textit{StorySync} achieves superior performance in comparison to SOTA training-free consistent subject generation approaches in both subject consistency and prompt adherence. Notably, \textit{StorySync} is model-agnostic and can be integrated with any state-of-the-art diffusion model without additional training. Our evaluations with SDXL, Kandinsky 3, and FLUX.1-schnell, demonstrate superior quality with these models. \textit{StorySync} enhances consistency of the generated subjects while also ensuring sufficient creative diversity.

\section{Acknowledgments}

\textit{We gratefully acknowledge Zebracat AI for providing the resources and support that made this research possible, and we thank the team for their valuable insights and contributions throughout the development of this work.}

{\small
\bibliographystyle{ieee_fullname}
\bibliography{storysync}
}


\clearpage
\appendix
\section{Appendices}

\subsection{Segmentation subject masks}\label{appendix:segmentation-masks}

\begin{figure}[!ht]
\centering
    \includegraphics[width=0.7\linewidth]{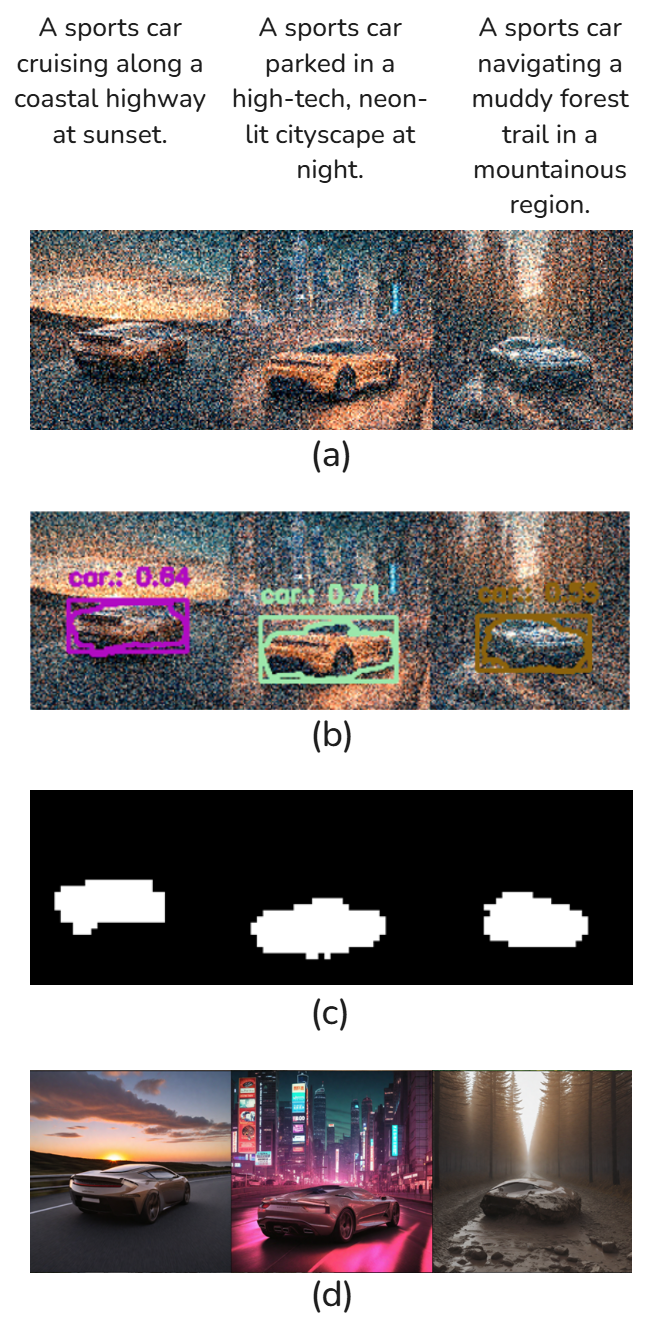}
    \caption[Problem with Segmentation masks]{\textbf{Problem with Segmentation masks} Using subject masks (c) generated using image segmentation (b) from intermediate noisy latents (a) generates visually poor subjects (d)}
    \label{fig:seg-mask}
\end{figure}

 Subject Masks generated using the cross-attention maps may sometimes be noisy and sometimes fail to capture subject-specific details, although this is a rare occurrence. We experiment with using segmentation techniques on intermediate noisy latents in the diffusion pipeline to generate the subject masks to be used for Cross-Image Attention Sharing and Regional Feature Harmonization. As seen in Figure \ref{fig:seg-mask}, to generate segmentation subject masks at a timestep \(t\), (1) we decode the latents to a noisy RGB image using the Variational Auto Encoder of the Diffusion Pipeline, (2) we pass the noisy RGB images through Grounding-DINO \cite{liu2024groundingdinomarryingdino} pipeline for subject identification in the noisy images, (3) we use Segment Anything Model \cite{kirillov2023segment} Vision Transformer (ViT) to create segmentation masks for the identified subjects in the images. These segmentation masks are then used as subject masks for Cross-Image Attention Sharing and BLI parts of our approach.

In Figure \ref{fig:seg-mask}, we can observe that the images generated after utilizing segmentation masks as the attention masks are of poor quality. We hypothesize that when we strictly localize the attention sharing among the subject patches only, we also prevent sharing important self-attention information that leads to generation of well-formed subjects. This observed behavior helps us identify an important aspect of image generation capability of the diffusion models, wherein the information required for proper formation of subjects is not localized only to the subject regions.

\begin{figure*}[!ht]
    \centering
    \includegraphics[width=0.8\linewidth]{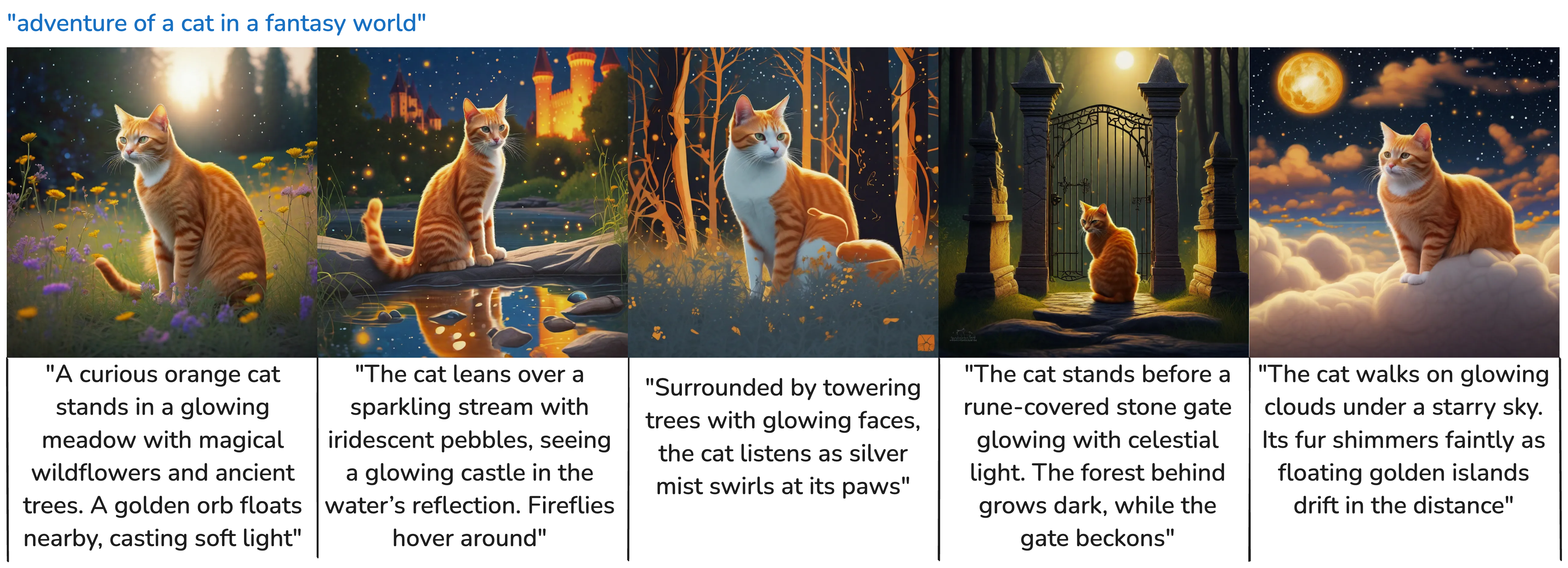}
    \caption[Subject consistency with complex prompts]{\textbf{Performance with complex prompts} Our approach is successfully able to generate visually similar subjects while also adhering to the complex requirements requested in a complex prompt}
    \label{fig:complex-prompts}
\end{figure*}

\begin{figure*}[!ht]
    \centering
    \includegraphics[width=0.8\linewidth]{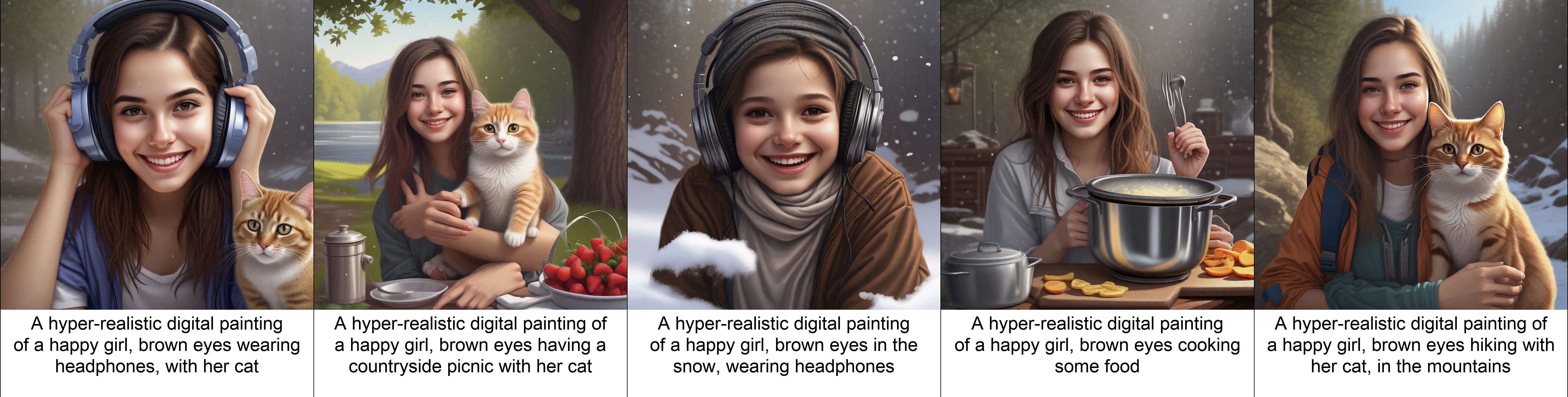}
    \caption[Multiple subjects generation]{\textbf{Consistency across multiple subjects} Even when multiple subjects are present in a batch of images, StorySync is able to generate visually similar multiple subjects}
    \label{fig:multi-subject}
\end{figure*}

\subsection{Human Evaluation}\label{appendix:human-evaluation}

To assess the quality of consistent subject generation while maintaining prompt adherence, we conducted a comprehensive human evaluation study. Human assessment serves as the gold standard for evaluating image generation quality, as it captures perceptual nuances that automated metrics may miss. We recruited ten human raters to evaluate image quality across three methods: \emph{ConsiStory}, \emph{StoryDiffusion}, and our proposed approach. Each rater assessed 25 randomly selected sets, with each set containing 5 images generated by the respective methods. The evaluation focused on two key criteria: (1) Subject Similarity, measuring the consistency of character appearance across images within a set, and (2) Prompt Adherence, evaluating how well the generated images align with the given textual prompts.

\begin{table}
\begin{tabular}{lcc}
\hline
Approach & Subject Sim. $\uparrow$ & Prompt Adh. $\uparrow$ \\
\hline
ConsiStory & 0.321 & 0.271 \\
StoryDiffusion  & 0.335 & 0.357 \\
\textit{StorySync (Ours)} & \textbf{0.344} & \textbf{0.372} \\
\hline
\\
\end{tabular}
\caption{\textbf{Human Evaluation}: Comparison of subject consistency and prompt adherence across different methods. Scores represent session-averaged preference votes, with higher values indicating better performance. Our method achieves superior performance in both subject similarity and prompt adherence metrics.}
\label{tab:user-study}
\end{table}

Table~\ref{tab:user-study} presents the human evaluation results. Our method demonstrates superior performance compared to baseline approaches across both evaluation metrics. The results indicate that human evaluators consistently preferred our generated images, confirming the effectiveness of our approach in balancing subject consistency with prompt fidelity. The human evaluation results validate that our approach successfully addresses the fundamental challenge of maintaining character consistency across diverse narrative scenarios.

\subsection{Complex prompts and multiple subjects}\label{appendix:complex-prompts-multiple-subjects}

Training-free approaches that interfere with the model's attention architecture are prone to interfering with the image generation capabilities of the model which can deteriorate the model's capability to follow the instructions provided by fairly complex prompts accurately. We test StorySync to generate consistent subjects while also adding complex background and environment details to be generated along with the subject. We observe in Figure \ref{fig:complex-prompts}, that StorySync not only generates almost similar subjects in these images but it is also successful in generating the intricacies of the background environment as outlined in the input prompts.

We can also observe in Figure \ref{fig:multi-subject}, that StorySync is able to generate multiple consistent subjects across a batch of images. For example, in the Figure \ref{fig:multi-subject}, in the images 1, 2 and 5, we can see that the two subjects (\textit{girl}, and \textit{cat}), are visually similar in these images.

\subsection{Quantitative Ablation Study}\label{appendix:quant-ablation-study}

\begin{table}[!ht]
\centering

\begin{adjustbox}{width=1.0\linewidth}

\begin{tabular}{lcccc}
\hline
Config & CLIP-T $\uparrow$ & CLIP‑I $\uparrow$ & LPIPS$\uparrow$ & DreamSim$\downarrow$ \\
\hline
Base & 0.8749 & 0.7819 & 0.3497 & 0.5263 \\
No Mask  & 0.7719 & 0.8812 & 0.4227 & \textbf{0.2782} \\
No RFH & \textbf{0.8125} & 0.8641 & 0.4022 & 0.2937 \\
No PV & 0.7918 & \textbf{0.8851} & \textbf{0.4198} & 0.2933 \\
\hline
Niblack & 0.8001 & 0.8729 & 0.4012 & 0.2885 \\
Adaptive & 0.8057 & 0.8664 & 0.3992 & 0.2913 \\
Sauvola & 0.8102 & 0.8686 & 0.4115 & 0.2905 \\
\hline
Full (\underline{Otsu}) & 0.8108 & 0.8735 & 0.4143 & 0.2869 \\
\hline
\\
\end{tabular}
\end{adjustbox}

(a)
\begin{adjustbox}{width=1.0\linewidth}
\begin{tabular}{llcccc}
\\
\hline
\multicolumn{2}{c}{Parameter} & CLIP-T $\uparrow$ & CLIP‑I $\uparrow$ & LPIPS$\uparrow$ & DreamSim$\downarrow$ \\
\hline
\multirow{3}{*}{$\gamma$} & \underline{0.3} & \textbf{0.8108} & 0.8735 & 0.4143 & 0.2869 \\
 & 0.5 & 0.7964 & 0.8763 & 0.4181 & 0.2813 \\
& 0.7 & 0.7699 & \textbf{0.8798} & \textbf{0.4214} & \textbf{0.2785} \\
\hline
\multirow{3}{*}{$\lambda$} & 0.3 & 0.7947 & \textbf{0.8782} & \textbf{0.4181} & \textbf{0.2844} \\
 & 0.5 & 0.7995 & 0.8768 & 0.4166 & 0.2855 \\
 & \underline{0.7} & \textbf{0.8108} & 0.8735 & 0.4143 & 0.2869 \\
\hline
\\
\end{tabular}
\end{adjustbox}

(b)

\vspace{1em}

\caption{\textbf{Ablation}: (a) Scores achieved for different components, and (b) $\gamma$ and $\lambda$ values. Parameter values used in StorySync are \underline{underlined} and optimal metric values are \textbf{bold}.}
\label{tab:ablation}
\end{table}

In addition to the qualitative ablation study, we also perform a quantitative ablation study and present the results in Table \ref{tab:ablation}. To perform this study, we use the dataset we originally used for quantitative analysis of our approach and generate and evaluate the images while disabling a component of StorySync with each run, and on similar evaluation metrics (CLIP-T for prompt adherence, and CLIP-I, LPIPS, and DreamSim for subject similarity). In Table \ref{tab:ablation}a, we compare the effect of disabling each component of our approach (subject masks, RFH, pose variation) and using different mask thresholding techniques such as Niblack, Sauvola, Adaptive, and Otsu. We can observe that when we do not use attention masks, the prompt adherence decreases; however, the perceptual similarity of the images increases because more parts of images are similar to each other now that masks do not block attention sharing. On the other hand, if we disable RFH, the prompt adherence increases because RFH enforces subject similarity and slightly impacts the prompt adherence of the generated images; however, that is a trade-off for generating visually similar subjects. If we disable pose variation techniques such as BLI and dropouts and generate the images, this leads to an increase in perceptual similarity of the images again; however, it reduces prompt adherence now that we are forcing the images to be more similar by limiting their layouts. 

In Table \ref{tab:ablation}b, we observe the effects of $\gamma$ and $\lambda$ as their values are ranged from 0.3 to 0.7 with an interval of 0.2. We observe that lower values of $\gamma$ allow for images with significant prompt adherence with slightly higher perceptual similarity, and hence in our experiments we fix the value of $\gamma$ to be 0.3. On the other hand, we observe that higher values of $\lambda$ lead to a significant increase in prompt adherence, and hence its value is fixed at 0.7 in our experiments.

\end{document}